\documentclass[10pt,conference]{IEEEtran}
\usepackage{textpos}
\usepackage{cite}
\usepackage{amsmath,amssymb,amsfonts}
\usepackage{algorithmic}
\usepackage{graphicx}
\usepackage{textcomp}
\usepackage{xcolor}
\def\BibTeX{{\rm B\kern-.05em{\sc i\kern-.025em b}\kern-.08em
    T\kern-.1667em\lower.7ex\hbox{E}\kern-.125emX}}
\usepackage{graphicx}

\usepackage{pgfplots}
\usepackage{pgfplotstable}
\usepackage{booktabs, colortbl, array}
\usepackage{enumitem}
\usepackage{rotating}
\usepackage{balance}
\usepackage{bbm}
\usepackage{amssymb}
\usepackage{amsmath}
\usepackage{amsthm}

\newtheorem{definition}{Definition}
\usepackage{url}

\begin{document}
\title{Massively Parallel Graph Drawing and Representation Learning}

\author{\IEEEauthorblockN{Christian B\"{o}hm}
\IEEEauthorblockA{\textit{Institute of Informatics}\\ \textit{Ludwig-Maximilians-Universit\"{a}t}\\
Munich, Germany\\
boehm@ifi.lmu.de}
\and
\IEEEauthorblockN{Claudia Plant}
\IEEEauthorblockA{\textit{Faculty of Computer Science, ds:UniVie}\\ \textit{University of Vienna}\\
Vienna, Austria\\
claudia.plant@univie.ac.at}
}

\maketitle

\begin{textblock*}{100mm}(0cm,-7.2cm)
\noindent Extended version of our paper at IEEE BigData 2020\cite{mpgempe}.
\end{textblock*}
%
\begin{abstract}
To fully exploit the performance potential of modern multi-core processors, machine learning and data mining algorithms for big data must be parallelized in multiple ways. Today's CPUs consist of multiple cores, each following an independent thread of control, and each equipped with multiple arithmetic units which can perform the same operation on a vector of multiple data objects. Graph embedding, i.e. converting the vertices of a graph into numerical vectors is a data mining task of high importance and is useful for graph drawing (low-dimensional vectors) and graph representation learning (high-dimensional vectors). In this paper, we propose MulticoreGEMPE (Graph Embedding by Minimizing the Predictive Entropy), an information-theoretic method which can generate low and high-dimensional vectors. MulticoreGEMPE applies MIMD (Multiple Instructions Multiple Data, using OpenMP) and SIMD (Single Instructions Multiple Data, using AVX-512) parallelism. We propose general ideas applicable in other graph-based algorithms like \emph{vectorized hashing} and \emph{vectorized reduction}. Our experimental evaluation demonstrates the superiority of our approach.
\end{abstract}
\section{Introduction}\label{sec:intro}
\noindent Today's processors are typically composed of multiple \emph{cores} which can process parallel tasks in a ``Multiple Instructions Multiple Data'' (MIMD) fashion, i.e. each core follows an independent thread of execution, allowing independent case distinctions (if-else), loops, and recursion (using independent stacks). These cores in principle share the same main memory although cache memories at various levels and new hardware architectures (like NUMA, non-uniform memory accesses) may result in varying cost (delays) of memory accesses depending on how local these accesses are. Nevertheless, all parallel tasks have direct access to all memory.

Each of the independent cores is composed of multiple arithmetic units which can process the same operation in parallel on multiple data in a ``Single Instructions Multiple Data'' (SIMD) way, e.g. add two vectors of real numbers, each composed from e.g. 16 components. Today, many processors from the well-known vendors support an instruction set extension called AVX-512 (Advanced Vector Extensions for 512 bit vectors). A 512 bit vector may comprise up to 16 int or float values of single precision (32 bit) or up to eight values of double precision (64 bit). In a single tact cycle, complete vectors may be added or multiplied or even both (using the ``fused multiply add''-operation, FMA), but also a large number of other mathematical, logical, load and store operations is possible. To handle vector objects, each core has a number of vector registers with fast access (32 vector registers for together up to 512 single-precision values in AVX-512).
\begin{figure}[t]
\centering
\vspace*{4mm}
\includegraphics[width=0.85\columnwidth]{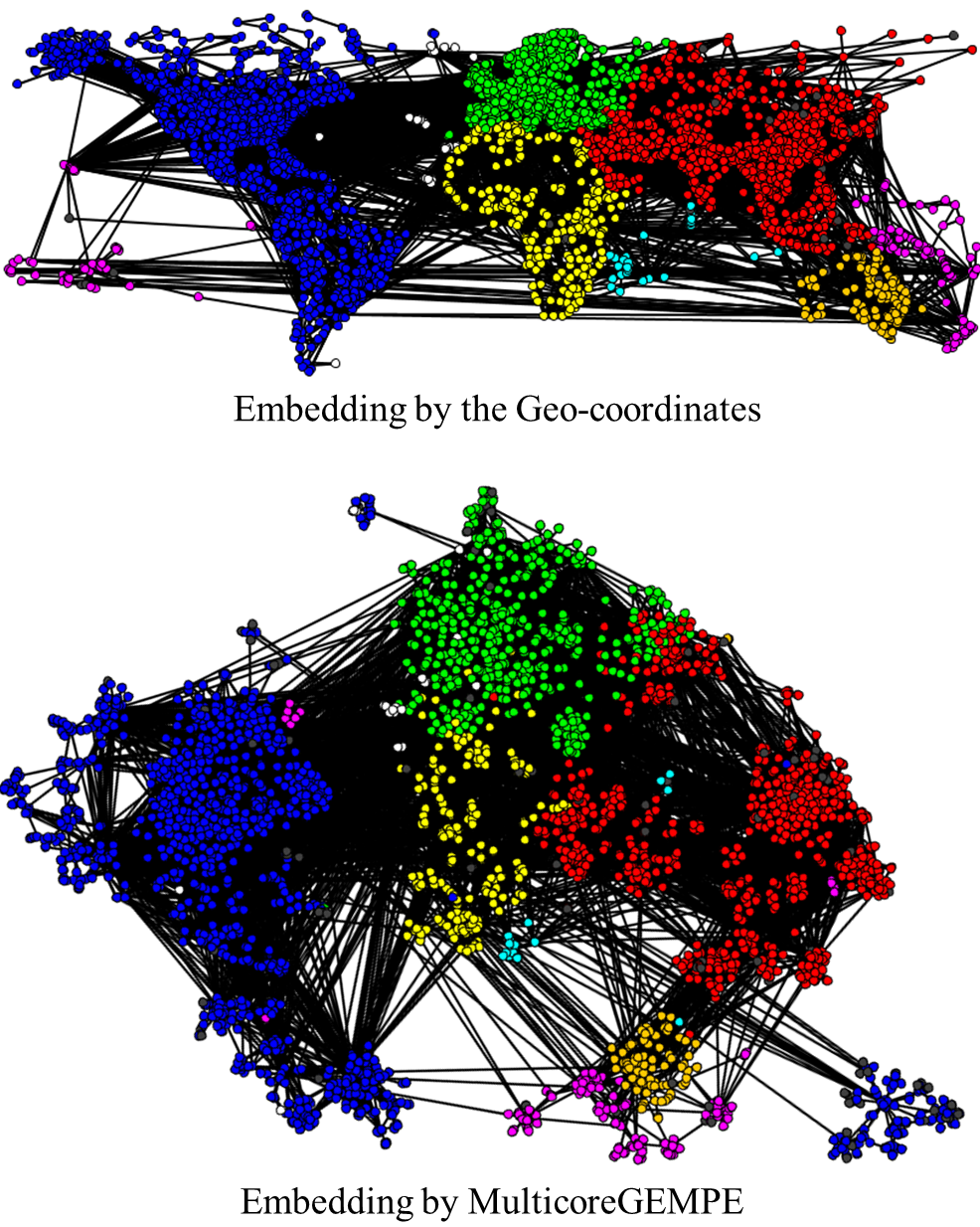}
\vspace{-2mm}
\caption{The World Flight Network.}
\label{fig:goodBad}
\vspace{-2mm}
\end{figure}
Since today not only super-computers but also very affordable commodity hardware may be able to process 128, 256 or even more operations in a single clock cycle, it is essential for modern algorithms to fully exploit their potential of SIMD and MIMD parallelism. Some well-known data analysis algorithms for big data have therefore been considered in this way: k-Means clustering \cite{DBLP:conf/sdm/BohmPP17,DBLP:conf/bigdataconf/BohmPP16}, Cholesky Decomposition \cite{IEEE:transbigdata/BohmPP18}, and Similarity Joins \cite{DBLP:conf/sigmod/PerdacherPB19}.

Graph Drawing and Graph Representation Learning are very important tasks of data mining on big data. Many applications produce graph-structured data, e.g., social, traffic, or protein networks. A graph with $n$ nodes is represented by a binary $n \hspace{-0.4mm}\times\hspace{-0.6mm} n$ adjacency matrix, or equivalently by an adjacency list. These representations are difficult to understand. Consider, e.g., the world flight network with over 3,000 nodes. It is much more accessible to view the graph in a 2D illustration with coordinates produced by a \emph{graph drawing technique}. However, in general it is difficult to assess the quality of graph drawing methods since usually no ground truth is available. For the world flight network, we have the geo-coordinates of each node corresponding to an airport. Figure \ref{fig:goodBad} compares the drawing of our new method MulticoreGEMPE (bottom) to the geo-coordinates. Our method preserves the basic layout of the continents and even the shape of them to some extent - only using the network of flight connections as an input.

Generating suitable vectors for the vertices of a graph is even more important in a second problem setting: Most techniques of machine learning and data mining like classification, clustering, anomaly detection, and regression work on vector data. Thousands of data analysis methods could be applied to graphs if we manage to translate graphs into vectors, typically in a medium to high dimensional space of $\ge\hspace{-0.6mm}100$ dimensions. This translation problem is referred to as \emph{graph representation learning}.

How to embed a graph in a low or high dimensional vector space? The research communities of graph drawing and representation learning have proposed sophisticated algorithms optimizing diverse goals (for more information see Section~\ref{sec:rw}). Only disproportionally few approaches to graph embedding support low- and high-dimensional vector spaces, where \cite{GEMPE} is one of the few exceptions supporting graph drawing and graph representation learning.

\textbf{MulticoreGEMPE considers graph embedding as a data compression task: The more effectively the low-dimensional coordinates allow to compress the information in the adjacency matrix, the better is the quality of the embedding.} Any kind of structure or patterns in graphs including clusters, hubs and spokes is expressed in the adjacency matrix. By faithfully preserving the information in the adjacency matrix we expose any kind of interesting structure to visualization and further data mining steps.

\subsection*{Contributions}
\begin{list}{}{\setlength{\leftmargin}{4.5mm}\setlength{\labelwidth}{4.5mm}\setlength{\labelsep}{1mm}}
\item[1.] We introduce MulticoreGEMPE (Graph Embedding by Minimizing the Predictive Entropy), a novel graph embedding method which is particularly suited for modern multi-core processors.
\item[2.] MulticoreGEMPE optimizes an objective function called \emph{predictive entropy} ($PE$) using a \emph{weighted majorization} approach with \emph{negative sampling}.
\item[3.] MulticoreGEMPE uses vectorization (SIMD parallelism with AVX-512) where edges and vertices of a graph can be efficiently accessed using \emph{gather} and \emph{scatter} operations and edge tests can be performed using \emph{vectorized hashing}.
\item[4.] MulticoreGEMPE is also parallelized at the level of edges and sampled non-edges using OpenMP (MIMD parallelism) with an efficient explicit reduction mechanism collecting information from SIMD and MIMD parallel computations.
\end{list}

\section{Preliminaries on Graph Embedding}
\noindent In this section we introduce the fundamental notations of Graph Embedding, Graph Drawing, and Graph Representaion Learning particularly following \cite{GEMPE}.

\subsection{Formal Problem Statement of Graph Embedding}
\begin{definition}[Graph Embedding.]
We are given a graph $G=(V,E)$ with $n$ vertices $V=\{1, ..., n\}$ and $m$ (with $m\le N=\tfrac{1}{2}n\cdot(n-1)$) unlabeled, unweighted, and undirected edges $(i,j)\in E$. Given is also the dimensionality $d\in \mathbb N$. The task is to find for each vertex $i\in V$ a vector $\mathbf x_i\in \mathbb R^d$ such that an objective function $f(\mathbf x_1, ..., \mathbf x_n)$ is minimized. If the dimension is low ($d\le 3$) we call the task \emph {Graph Drawing}, otherwise \emph{Graph Representation Learning}.
\end{definition}

The most prevalent setting for graph drawing is $d=2$ and for graph representation learning $d=128$ or $d=256$.

\subsection{Predictive Entropy PE}
\noindent GEMPE defines an objective function for graph embedding called the \emph{Predictive Entropy} $(PE)$. The $PE$ function involves a probability function $s(||\mathbf x_i-\mathbf x_j||)$ predicting for any pair of vertices $i$ and $j$ the probability that an edge $(i,j)$ exists. Given a good embedding, the $PE$ function is able to predict all entries almost perfectly with probabilities close to zero for non-edges and close to one for edges. $PE$ represents the number of bits required to encode the remaining uncertainty on the edge existence given the low-dimensional coordinates. A perfect embedding has a $PE$ of 0 bits.


The basic representation of an unweighted, undirected graph is the upper triangle of the adjacency matrix which requires a number $N=\frac{1}{2}n\cdot(n-1)$ of bits, one bit for each pair of vertices. This graph representation can be efficiently and lossless compressed by entropy-based methods like Huffman coding or arithmetic coding which assign to each object a bit string of the length $-\log_2 P$ where $P$ is the (estimated) probability of the object. In a basic setting (without using our \emph{knowledge}), we assume for each vertex pair the same probability of $P_{\mbox{\scriptsize basic}}\big((i,j)\in E\big) = \frac{m}{N}$ to be an edge where $m=|E|$ is the actual number of edges in our graph. Since we have a description length of $-\log_2\frac{m}{N}$ for an edge and $-\log_2\frac{N-m}{N}$ for a non-edge, the average description length of the vertex pairs corresponds to the basic entropy of the adjacency matrix:
\begin{equation}
\mathrm H_{\mbox{\scriptsize basic}} = \mbox{$-\frac{m}{N}\log_2\frac{m}{N}-\frac{N-m}{N}\log_2\frac{N-m}{N}$},
\label{eq:basicEntropy}
\end{equation}
which is between $0$ and $1$, often $\ll 1$. $\mathrm H_{\mbox{\scriptsize basic}}$ is also a lower bound of the size of the run-length coded adjacency matrix, the adjacency list, and compressed versions thereof.

Our idea is that we can exploit any given embedding to get a better estimation of the probability that a vertex pair $(i,j)$ is an edge, and thus a better compression. It is useful to apply the Euclidean distance $||\mathbf x_i - \mathbf x_j||$ between the associated vectors of the embedding and, instead of the constant probability $P_{\mbox{\scriptsize basic}}\big((i,j)\in E\big) = m/N$ to consider the conditional probability, given this distance:
\[P\big((i,j)\in E\mbox{ }\big|\mbox{ }||\mathbf x_i - \mathbf x_j|| = \delta\big).\]
In our method $PE$, we model this conditional probability by a sigmoid function analogous to the (negative) cumulative normal distribution which is defined using the error function:
    \[\hspace{-1mm}\begin{array}{r@{\hspace{1mm}:=\hspace{1mm}}c@{=}l}s^{+}_{\mu,\sigma}(\delta) & P\big((i,j)\in E\mbox{ }\big|\mbox{ }||\mathbf x_i - \mathbf x_j|| & \delta\big) =  \mbox{$\frac{1}{2}-\frac{1}{2} \hspace{0.5mm}$erf$\big(\frac{\delta-\mu}{\sqrt 2\sigma}\big)$},\vspace{0.5mm}\\
    s^{-}_{\mu,\sigma}(\delta) & P\big((i,j)\notin E\mbox{ }\big|\mbox{ }||\mathbf x_i - \mathbf x_j|| & \delta\big) =  \mbox{$\frac{1}{2}+\frac{1}{2}\hspace{0.5mm} $erf$\big(\frac{\delta-\mu}{\sqrt 2\sigma}\big)$}.\end{array}\]
Throughout this paper we will use superscripts to distinguish between the two cases \emph{edge} $(^{+})$ and \emph{non-edge} $(^{-})$.

\begin{figure}
\centering
\includegraphics[width=1.0\columnwidth]{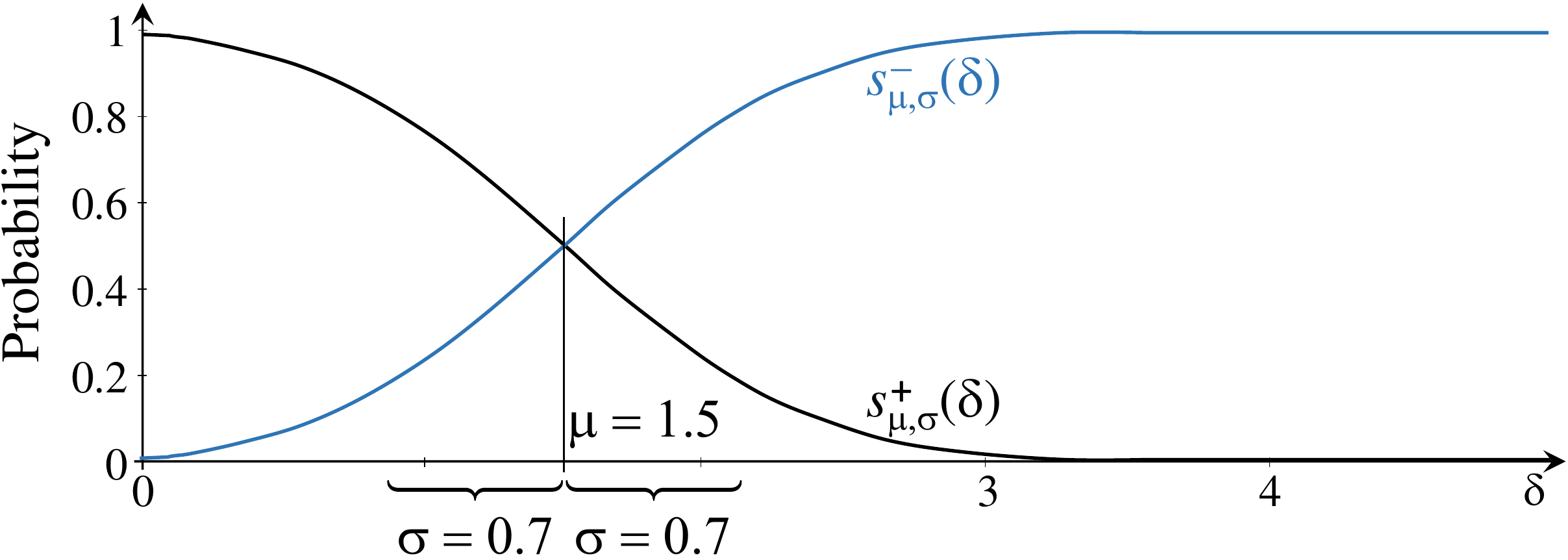}
\vspace{-2mm}
\caption{The Edge Existence Probability $s^{+}_{\mu,\sigma}(\delta)$ and its Opposite $s^{-}_{\mu,\sigma}(\delta)$ .}
\label{fig:parabola}
\end{figure}

The sigmoid function $s^{+}_{\mu,\sigma}(\delta)$ is depicted in Figure~\ref{fig:parabola}. It combines several desirable properties, namely it is strictly monotonic increasing from a plateau of values close to 1 to a plateau of values close to zero, with an (optimizable) slope in between.

Since $s^{+}_{\mu,\sigma}(\delta)$ models the probability that two vertices are connected by an edge, the \emph{description length dl} (the number of bits required by the $(i,j)$-entry of the adjacency matrix in an entropy-based coding) corresponds to the negative binary logarithm of $s^{+}_{\mu,\sigma}(\delta)$ or $s^{-}_{\mu,\sigma}(\delta)$:
\[\begin{array}{l@{\hspace{2mm}=\hspace{2mm}}l}
\mbox{\textit{dl}}^{+}_{\mu,\sigma}(\delta) & \mbox{$-\log_2\big(\frac{1}{2}-\frac{1}{2}\hspace{0.5mm} \mbox{erf}(\frac{\delta-\mu}{\sqrt 2\cdot \sigma})\big)$}\vspace{0.8mm}\\
\mbox{\textit{dl}}^{-}_{\mu,\sigma}(\delta) & \mbox{$-\log_2\big(\frac{1}{2}+\frac{1}{2}\hspace{0.5mm} \mbox{erf}(\frac{\delta-\mu}{\sqrt 2\cdot \sigma})\big)$}\vspace{0.8mm}\\
\mbox{\textit{dl}}_{\mu,\sigma}(i,j) &  \left\{\hspace{-1mm}\begin{array}{ll} \mbox{\textit{dl}}^{+}_{\mu,\sigma}(||\mathbf x_i - \mathbf x_j||)&\hspace{-2mm}\mbox{if }(i,j)\in E\vspace{0.5mm}\\
\mbox{\textit{dl}}^{-}_{\mu,\sigma}(||\mathbf x_i - \mathbf x_j||)&\hspace{-2mm}\mbox{otherwise.}\end{array}\right.
\end{array}\vspace{0mm}\]
Intuitively, \emph{dl} punishes an entry $(i,j)$ of the adjacency matrix with a long bit string if (1) it is an edge but $||\mathbf x_i-\mathbf x_j||$ is large ($\gg \mu$), or (2) if $(i,j)\notin E$ but  $||\mathbf x_i-\mathbf x_j||$ is small ($\ll \mu$). In contrast, if we have a beautifully drawn graph with only edges having a distance $\ll \mu$ and non-edges $\gg\mu$ then \emph{dl} awards us with a $dl$ very close to $0$.

For an undirected graph, the adjacency matrix is symmetric, and only the upper triangle is coded. The \emph{predictive entropy} $PE$ of the graph is the average number of bits needed for each entry of the adjacency matrix after compression:
\begin{equation}
    PE_G(\mathbf x_1, ..., \mathbf x_n) = \mbox{$\frac{1}{N}$}\cdot\min_{\mu,\sigma}\sum_{1\le i\le n}\hspace{2mm}\sum_{i<j\le n} \mbox{\textit{dl}}_{\mu,\sigma}(i,j).
    \label{eq:costG}
\end{equation}
The function $PE_G(\mathbf x_1, ..., \mathbf x_n)$ can be used to assess the result quality of an arbitrary graph embedding method, because it measures how well we can predict an edge from the distance of the corresponding nodes in the embedding. The parameters $\mu$ and $\sigma$ are optimized in order to minimize $PE_G(\mathbf x_1, ..., \mathbf x_n)$. If any sigmoid exists which allows on average over all node pairs a good prediction of the edge existence, we have a good embedding, no matter what the values of $\mu$ and $\sigma$ actually are. In general, if $\sigma$ is small compared to $\mu$, then $\mbox{\textit{dl}}_{\mu,\sigma}(\delta)$ is very decisive, which tends towards small coding cost (provided that there are not many prediction errors of vertex pairs $\in E$ having a distance $>\mu$ or vertex pairs $\notin E$ having a distance $<\mu$).
\subsection{Weighted Majorization for PE Minimization}
\label{sec:gempe}
\noindent Methods related to multidimensional scaling (MDS) enable us to automatically determine the coordinates $\mathbf x_i$ for objects of which the matrix of all pairwise distances $d_{i,j}$ is known. MDS based on Eigenvalue decomposition requires the distance function to be a metric (symmetric, positive definite, and observing the triangle inequality), but more advanced MDS methods, in particular Weighted Majorization (WM) \cite{BrandesP08}, \cite{DBLP:conf/gd/GansnerKN04} support non-metric distance functions.
Moreover, WM allows us to apply a weight $w_{i,j}$ to each object pair specifying to which extent the corresponding distance is taken into account in the overall objective function. The technique aims at finding low-dimensional coordinates $\mathbf x_i$ that minimize the weighted squared difference between the distance in the input matrix $d_{i,j}$ and the Euclidean distance $||\mathbf x_i-\mathbf x_j||$ of the generated vectors:
\[wm(\mathbf x_1, ... , \mathbf x_n) = \sum_{1\le i\le n}\hspace{2mm}\sum_{i<j\le n}w_{i,j}\cdot(||\mathbf x_i-\mathbf x_j||-d_{i,j})^2.\]

Our coding function
 $\mbox{\textit{dl}}_{\mu,\sigma}(i,j)$ does not directly provide a matrix of pairwise distances $d_{i,j}$. However, since $wm(\mathbf x_1, ... , \mathbf x_n)$ optimizes for a squared error (a parabolic function), we can derive from $\mbox{\textit{dl}}_{\mu,\sigma}(i,j)$ a parabola $w_{i,j}\cdot (||\mathbf x_i - \mathbf x_j||-d_{i,j})^2$ which tightly approximates the in- or decrease of true cost. We require that for the current distance $\delta=||\mathbf x_i - \mathbf x_j||$, the description length function and the parabola agree in their \emph{first derivative} $\frac{\mathrm d}{\mathrm d\delta}$ and \emph{second derivative} $\frac{\mathrm d^2}{\mathrm d\delta^2}$ w.r.t distance $\delta$. For $(i,j)\in E$ we obtain:
\begin{eqnarray*}
    \mbox{$\frac{\mathrm d}{\mathrm d \delta}$}\mbox{\textit{dl}}^{+}_{\mu,\sigma}(\delta) & = &  \mbox{$\frac{\mathrm d}{\mathrm d \delta}$}w_{i,j}\cdot (\delta-d_{i,j})^2,\\
    \mbox{$\frac{\mathrm d^2}{\mathrm d \delta^2}$}\mbox{\textit{dl}}^{+}_{\mu,\sigma}(\delta) & = &  \mbox{$\frac{\mathrm d^2}{\mathrm d \delta^2}$}w_{i,j}\cdot (\delta-d_{i,j})^2.
\end{eqnarray*}
From these two equations, we determine $d_{i,j}$ and $w_{i,j}$, calling it from now on $(d^{+}_{i,j},w^{+}_{i,j})$ and obtain, after a few algebraic transformations: 
\begin{eqnarray}
\hspace*{-4mm} w_{i,j}^{\color{black}+\color{black}} \hspace{-2mm}& = \hspace{-2mm}&\frac{\frac{2}{\pi \ln 2}\big(\textrm e^{-\frac{(\delta-\mu)^2}{2\sigma^2}}\big)^2}{\sigma^2(1\color{red}-\color{black}\mbox{erf}(\frac{\delta-\mu}{\sqrt 2\sigma}))^2}
\color{red}-\color{black}
\frac{\frac{1}{\sqrt{2\pi}\ln 2}(\delta-\mu)\textrm e^{-\frac{(\delta-\mu)^2}{2\sigma^2}}}{\sigma^3(1\color{red}-\color{black}\mbox{erf}(\frac{\delta-\mu}{\sqrt 2\sigma}))}\label{eq:wplus}\\
\hspace*{-4mm}d_{i,j}^{\color{black}+\color{black}} \hspace{-2mm}& = \hspace{-2mm}& \delta\color{red}-\color{black}\frac{\frac{1}{\sqrt{2\pi}\ln 2}\textrm e^{-\frac{(\delta-\mu)^2}{2\sigma^2}}}{w_{i,j}\cdot \sigma(1\color{red}-\color{black}\mbox{erf}(\frac{\delta_{i,j}-\mu}{\sqrt 2\sigma}))}.\label{eq:dplus}
\end{eqnarray}
The corresponding formulas $(d_{i,j}^-,w_{i,j}^-)$ for the case $(i,j)\notin E$ are very similar but with the ``\color{red}$-$\color{black}'' operations (marked in red) changed into ``$+$''. In the general case we write $(d_{i,j},w_{i,j})$ meaning $(d_{i,j}^+,w_{i,j}^+)$ if $(i,j)\in E; (d_{i,j}^-,w_{i,j}^-)$ otherwise.
Since two derivatives agree, the actual \textit{dl} is very well approximated by the parabola for a wide range of values around the current distance $\delta$. Therefore, we can move $\mathbf x_i$ and/or $\mathbf x_j$ over a wide area in the next iteration without losing the good approximation. The parabola with its parameters $d_{i,j}$ and $w_{i,j}$ is used to determine the next set of coordinates in the next iteration.

As indicated before, GEMPE uses Weighted Majorization \cite{DBLP:conf/gd/GansnerKN04} for the computation of the next set of vectors given the parameters of the parabolas $(d_{i,j},w_{i,j})$. This iterative algorithm sets the new value $\mathbf x_i$ to:
\begin{equation}
    \mathbf x_i = \frac{\mathbf y_i}{z_i} = \frac{\sum_{j\neq i}w_{i,j}\cdot(\mathbf x_j+s_{i,j}\cdot(\mathbf x_i-\mathbf x_j))}{\sum_{j\neq i}w_{i,j}}
    \label{eq:weighted}
\end{equation}
where $s_{i,j} = \left\{\begin{array}{ll}d_{i,j}/||\mathbf x_i - \mathbf x_j|| & \mbox{if }||\mathbf x_i - \mathbf x_j|| \neq 0\\0&\mbox{otherwise.}\end{array}\right.$

\noindent We define the shortcuts $\mathbf y_i$ for the \emph{numerator} and $z_i$ for the \emph{denominator} to facilitate later discussion.

For a fixed sigmoid setting $(\mu,\sigma)$, we can provide the following algorithm: perform a loop until convergence which involves (1) determination of $(d_{i,j},w_{i,j})$ for all pairs of nodes $(i,j)$, and (2) determination of a complete set of new coordinates $(\mathbf x_1, ..., \mathbf x_n)$ for all nodes according to Eq.~(\ref{eq:weighted}). The proof of convergence of WM can be found in \cite{DBLP:conf/gd/GansnerKN04}.
\section{MulticoreGEMPE}
\subsection{Vectorized Graph Processing}
\noindent SIMD parallelism (Single Instruction Multiple Data) has become prevalent through instruction set extensions like SSE or AVX. These original assembler instructions have been mapped to compilers of C++ and other languages without the need to program in assembler. C++-mapped instructions (like \textbf{\_mm512\_add\_sd} for adding two 16-component vectors) are called AVX-intrinsics. Special types like \textbf{\_\,\_m512i} for a 16-component vector of integers (with the usual 32 Bit each, like the C++-type \textbf{int}) or \textbf{\_\,\_m512} for a 16-component vector of single-precision floating-point values (with the usual 32 Bit each, like the C++-type \textbf{float}) can be used for local variables of a C++ program. The instruction set extension AVX-512 defines for each core 32 registers for vectors of 512 Bit each, and local variables of vector type like \textbf{\_\,\_m512} are automatically mapped to one of these registers by the compiler's optimizer.

Our whole idea of parallelization and vectorization is based on the actual edges $E$ of the graph: The actual edges $(i,j)\in E$ are distributed in a block-wise way among the different processor cores, and each of these stores, in a pair of vector registers $(RI, RJ)$, a number of subsequent edges to be processed simultaneously by SIMD parallelism:
\[RI=\langle i_0, i_1, ..., i_{15}\rangle; \hspace{5mm}RJ=\langle j_0, j_1, ..., j_{15}\rangle.\]
The corresponding information like the coordinates of the embedded nodes, distances between these nodes, and probabilities can then be fetched and computed via AVX vector arithmetics.
\subsection{Vertex Accesses through Gather and Scatter}
\noindent A sequence of contiguously stored edges can be easily transferred from and into the AVX-512 registers using load- and store-operations like: $RI = $\textbf{\_mm512\_load\_epi32}\,(\emph{address});

If we then want to load the information associated to the nodes $\langle i_1, i_2, ...\rangle$ in another AVX register (like e.g. the current 2d-coordinates $\langle x[i_1], x[i_2] ...\rangle$ and $\langle y[i_1], y[i_2] ...\rangle$, we need vectorized operations for array accesses. These are called \textbf{gather} (for fetching indexed data out of an array) and \textbf{scatter} for writing indexed data into an array. The $x$-coordinates of the vertices $\langle i_1, i_2, ...\rangle$ are e.g. read by the operation:\[RX = \mbox{\textbf{\_mm512\_i32gather\_ps\,}}(RI, \mbox{\emph{address}}, 4);\]
where 4 is the length of each array element in bytes and \emph{address} is the start address of the array containing the $x$-coordinates.
Special care must be taken to avoid concurrent writing into the same array element, i.e. to use \textbf{scatter} when some of the indices in $RI$ are actually the same (e.g. $i_3 = i_7$). We will elaborate on this issue in Section~\ref{sec:reduction}.
\subsection{Vectorized Negative Sampling Using Vectorized Hashing}
\noindent It has been shown for other graph-processing algorithms \cite{DBLP:conf/www/TangQWZYM15,GEMPE} that a strategy called ``\emph{negative sampling}'' does not only improve the run-time performance directly through the reduced, sampled-out workload but also increases the convergence speed, reduces the number of iterations, and improves the overall result quality. Negative sampling means in this context to consider all the true edges (the ``positives'') but to consider only a small sample of those vertex pairs $\notin E$. The strategy of MulticoreGEMPE is to generate for every true edge $(i,j)\in E$ two sample vertices $g, h\in V$ for which $(g,i)\notin E$ and $(h,j) \notin E$. This strategy implicitly pairs vertices of high degree with linear more non-edges, as proven optimal in \cite{DBLP:conf/www/TangQWZYM15}. Moreover, this strategy makes MulticoreGEMPE obviously linear in the number of edges (opposed to $O(n^2)$ for the non-sampling version).

For the generation of samples $g,h\in V$, we implemented a vectorized linear congruential random number generator (LCG) \cite{Lehm51,random} with 32 Bit similar to \textbf{rand()} in C++ which used a multiplier of $1103515245$ and an increment of $1$. Although not suitable for information security applications, this random number generator is highly efficient in runtime and fulfills the requirements of randomness of simulation applications, as well as of our negative sampling. Our vectorized LCG works as follows: An AVX-register $RS$ stores a vector of 16 state variables (of 32 Bit each), initialized with different random bit patterns. Whenever new random numbers have to be generated, these state variables are updated by multiplier and increment:
\[\begin{array}{l}RS = \mbox{\textbf{\_mm512\_add\_epi32}}\,(1, \\
\hspace{14mm}\mbox{\textbf{\_mm512\_mullox\_epi32}}\,(1103515245, RS));\end{array}\]
where \textbf{\_mm512\_mullox\_epi32} determines the lower 32 Bit of the result of the multiplication, as needed for an LCG. The values $1103515245$ and $1$ must be broadcasted in a register by \textbf{\_mm512\_set1\_epi32}, which is not shown here for readability. Then, depending on a mask (described in the next paragraph) the state variables is transformed into the range $\{0, ..., n-1\}$ using bitwise logical operations (\textbf{\_mm512\_and\_epi32}, \textbf{\_mm512\_or\_epi32}), a floating point multiplication, and rounding. After this step, we have a register $RG$ containing the nodes $\langle g_0,...,g_{15}\rangle$ and register $RH=\langle h_0,...,h_{15}\rangle$.

Whenever, for any of the nodes in $RG$, the corresponding node in $RI$ is an actual edge (the test is described in the next paragraph), the sampling must be repeated for those pairs which are actual edges. This is controlled by the mask mentioned in the last paragraph which prevents to replace random nodes which are non-edges already. Note that it might be unlikely to randomly sample a complete vector of nodes with non-edges, even if the sampling is often repeated. Replacing only those vertices with actual edges requires considerable fewer tries. The same applies for the non-edge node pairs in AVX-registers $RK$ and $RJ$.

The test whether or not a node pair is connected by an edge is done by a hashing method which stores at an address which is associated to a pair $(i,j)$ a byte value of $1$ if $(i,j)\in E$. If neither $(i,j)$ nor any node pair that has a hashing collision with $(i,j)$ has an edge, then $0$ is stored at the address. The size $s$ of the address space is chosen as a power of two to facilitate computation with bitwise logical operations and $\gg m$ (by some orders of magnitude) to make collisions rare. As hashing method for vertex pair $(i,j)$, we define:
\[\mbox{hash}\,(i,j) = ((i\cdot n + j)\color{red}*\color{black} 1103515245) ~\color{red} \& \color{black}~(s-1)\]
where ``$\color{red}*\color{black}$'' and is a 32-bit integer multiplication where overflow is ignored and only the lower 32 bit of the result are actually used. Similarly, ``$\color{red}\&\color{black}$'' is the bitwise ``\textbf{and}'' operation on 32 bit.

The whole algorithm of determining the non-edge partners $RG$ of the nodes in $RI$ is given in a C++-style pseudo-code in Figure~\ref{fig:nonedges}. This is computed after the vertices have been renamed randomly and this renaming can be repeated after a certain number of iterations to prevent hashing collisions from causing systematic embedding errors. The effort for this is negligible.

\begin{figure*}[t]
  \centering
  \fbox{
    \begin{minipage}{0.975\textwidth}
      \fontsize{7}{8}\selectfont
      \hspace*{-0.5em}\textbf{algorithm} vectorizedSampleNonEdges (\textbf{\_\,\_m512i} $RI$)\,\{\\
      \hspace*{0.2em}\textbf{\_\,\_m512i} $RG = 0;$\hspace{5mm}\textbf{\_\,\_m512} $RN = $\textbf{\_mm512\_set1\_ps}\,$((\mbox{\textbf{float}})\,\,n);$\hspace{5mm}\textbf{int} \emph{mask} = 0xffff; \hspace{49mm}// all $16$ bits of \emph{mask} set to `$1$'\hspace*{-1mm}\\
      \hspace*{0.2em}\textbf{while} (\emph{mask} $\neq 0$)\,\,\{\hspace{80mm}// repeat as long as there are edge pairs in $RI$ and $RG$, e.g. $(i_3, g_3)\in E$\hspace*{-1mm}\\
      \hspace*{0.9em}$RS \hspace{1.22mm} = \mbox{\textbf{\_mm512\_add\_epi32}}\,(1, \mbox{\textbf{\_mm512\_mullox\_epi32}}\,(\mbox{1103515245}, RS));$\hspace{37mm}// update the state vector of the random number generator\hspace*{-1mm}\\
      \hspace*{0.9em}$RG \hspace{0.92mm} = \mbox{\textbf{\_mm512\_mask\_cvt\_roundps\_epi32}}\,(RG, \mbox{\emph{mask}}, \mbox{\textbf{\_mm512\_or\_epi32}}\,(\mbox{0x3f800000}, \mbox{\textbf{\_mm512\_and\_epi32}}\,(RS, \mbox{0x7fffff}))* RN - RN)$;\vspace{-0mm}\\
      \hspace*{0.9em}\emph{mask} $=$ \textbf{\_mm512\_cmpneq\_epi32\_mask}\,($0$,\,\textbf{\_mm512\_i32gather\_epi32}\,(\textbf{\_mm512\_mullo\_epi32}\,(\textbf{\_mm512\_mullo\_epi32}\,(\emph{$RI$,\,$n$})$\mbox{}+ RG$,\,1103515245)\,\&\,($s\hspace{-0.5mm}-\hspace{-1mm}1$), \emph{hmap}, $1$)\,\,\&\,$1$);\hspace*{-1mm}\\
      \hspace*{-0.5em}\}\hspace*{0.2em}\}
  \end{minipage}
  }
  \caption{The Algorithm for Vectorized Non-edge Sampling (some typecast and broadcast operations left out for readability).}
  \label{fig:nonedges}
  \vspace{-2mm}
\end{figure*}

\subsection{Vectorized exp- and erf-Approximations}
\noindent Most of the operations required to implement Eq.~(\ref{eq:dplus}) and Eq.~(\ref{eq:wplus}) are additions, subtractions, multiplications etc. and thus straightforward in AVX. However, the two functions exp and erf are not implemented as assembler instructions. For maximum speed we implemented them in an approximate way as piecewise functions, composed from 16 independent
quadratic functions (i.e. $ax^2+bx+c$) which are continuous at the connections. The 3 parameter vectors ($a,b,c$) are stored in AVX registers, and the valid parameter setting for a given $x$ is selected using the AVX-intrinsic \textbf{\_mm512\_permutexvar\_ps}. The case distinction of the weighted majorization step in Eq.~(\ref{eq:weighted}) can be implemented using masked operations like those in Fig.~\ref{fig:nonedges}.
\subsection{Explicit Vectorized Result Reduction}
\label{sec:reduction} \noindent A round of weighted majorization is (MIMD-) parallelized using OpenMP by assigning a contiguous block of $\approx m/t$ edges to each of the $t$ threads. Each thread then collects the changes to the numerator $\mathbf y_i$ (which is a $d$-dimensional vector) and denominator $z_i$ (which is a scalar value) of Eq.~(\ref{eq:weighted}) for each vertex $i$. Both numerator and denominator of Eq.~(\ref{eq:weighted}) are computed as a sum of some values. Therefore, numerator and denominator can be consolidated by collecting and adding up all information of the parallel threads. OpenMP already supports the ``\emph{reduce}''-clause to automatically add up such information, if the arrays containing $\mathbf y_i$ and $z_i$ are declared in a ``\emph{reduce}''-clause.

However, we have a related, similar problem in vectorization. As mentioned earlier, when calling a scatter operation, we have to ensure that no two indices have a collision, i.e. write to the same array element. Although some instruction set extensions (but not all those implementing AVX-512) support an explicit collision detection, we propose here a different solution that we call \emph{explicit vectorized result reduction}. In explicit vectorized result reduction, we define a separate version of numerator and denominator not only for each independent thread but also for each of the 16 components of the AVX-registers. By writing to the different array versions we automatically ensure no collisions. This is achieved by adding an AVX-register containing the row $\langle 0,n,2 n, 3 n, ...,15 n\rangle$ to the index vector $RI$ before calling scatter to fill, e.g. the array $z$ with values collected in register $RZ$ (the broadcast of $n$ into a register is again left out for readability):
\[\begin{array}{l}\mbox{\textbf{int} } \mbox{\emph{row}} = \{0, 1, 2,3,4,5,6,7,8,9,10,11,12,13,14, 15\};\\
\mbox{\textbf{\_\,\_m512i} } ROW = \mbox{\textbf{\_mm512\_loadu\_si512}}\,(\mbox{\emph{row}});\\
\mbox{\textbf{\_mm512\_i32scatter\_ps}}\,(\mbox{\textit{array}}, RI+n*ROW, RZ, 4);
\end{array}\]

\noindent Of course, the \textit{array} must be allocated with size $16 t\cdot n$ rather than $n$ when applying explicit vectorized result reduction. Before a thread ends (after every iteration of the overall weighted majorization), it \emph{reduces} its $16$ versions of the array into one by summing up all $16$ versions, again using the AVX intrinsic \textbf{\_mm512\_add\_ps}. Finally, the versions of the $t$ independent OpenMP threads are reduced.

\subsection{Histogram-based Slope Optimization}\label{sec:slope}
\begin{figure}[b]
\centering
\vspace{-2mm}
\includegraphics[width=0.90\columnwidth]{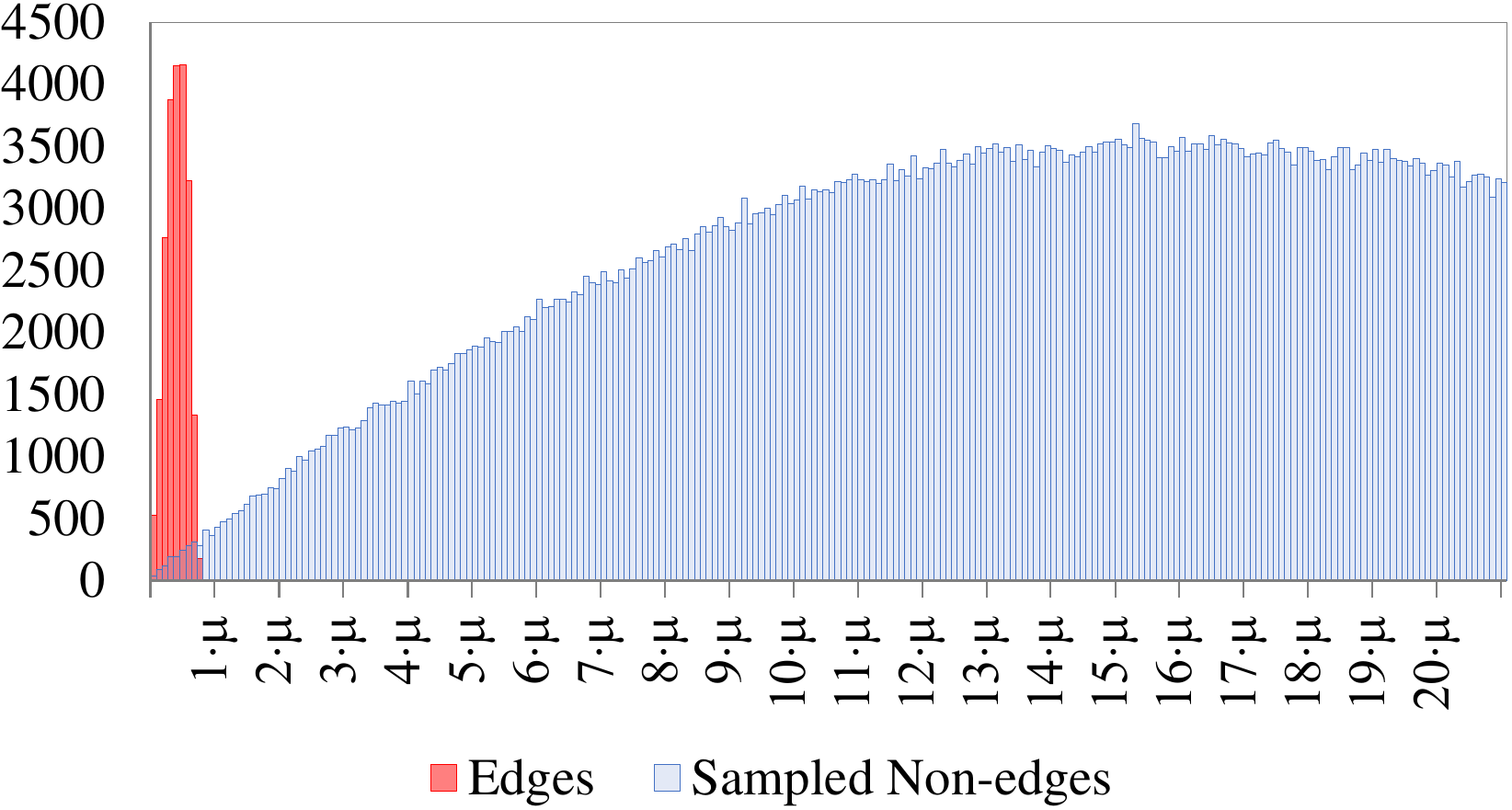}
\vspace{-1.5mm}
\caption{The Histogram of Distances}
\label{fig:histogram}
\end{figure}

\noindent While we fix the position parameter of the sigmoid $\mu=1.5$, the slope parameter $\sigma$ is optimized after every iteration of weighted majorization or after a fixed number of iterations. In principle, we do this by determining the derivative of Eq.~(\ref{eq:costG}),
\[\frac{\mbox{d}}{\mbox{d}\sigma}\,PE_G(\mathbf x_1, ..., \mathbf x_n) = 0.\]
While the derivative has an analytic solution, the (only) zero position is determined by a bisection search. It would be costly to perform this bisection search on the set of actual edges or (even worse) overall node pairs. Instead, we summarize the occurring distances in a histogram, like that in Fig.~\ref{fig:histogram}, separately for the edges and the non-edges. This histogram can either be collected during weighted majorization, when distances between node pairs are determined anyway. This comes at almost no extra cost but we use actually the distances after the previous iteration. It is also possible to determine the histogram for all edges and an arbitrary sample of the non-edges in a separate step parallelized similarly to the weighted majorization step, i.e. using OpenMP and AVX-512.

Then the bisection search for the optimal slope $\sigma$ is performed on the histogram, taking into account the different weights of the histogram bins of edges and non-edges if negative sampling was applied for the histogram. This step is inexpensive and needs no parallelization.

\subsection{The Overall Algorithm MulticoreGEMPE}
\begin{figure}[b]
  \centering
  \vspace*{-2mm}
  \fbox{
    \begin{minipage}{0.95\columnwidth}
      \fontsize{8}{9}\selectfont
      \textbf{algorithm} MulticoreGEMPE $(V, E, d)\rightarrow \mathbb R^{n\times d}$\vspace{0mm}\\
      \hspace*{1em}initialize $\mathbf x_0, ..., \mathbf x_{n-1}$ randomly from a uniform distribution; $\sigma:= 1$;\\
      \hspace*{1em}\textbf{repeat until convergence}\vspace{0mm}\\
      \hspace*{2em}initialize numerators $\mathbf y_i$ and denominators $z_i$ with $0$;\\
      \hspace*{2em}distribute $E$ among $t$ threads $\theta$ of size $m_\theta = 16\cdot\big\lfloor\frac{m}{16 t}\big\rfloor$;\\ \hspace*{2em}\textbf{foreach thread} $\theta \in \{0,...,t-1\}$ \textbf{do in parallel}\\
      \hspace*{3em}\textbf{for} $k:=0$ \textbf{to} $m_\theta-1$ \textbf{stepsize} $16$ \textbf{do}\\
      \hspace*{4em}$RI:=$ \textbf{\_mm512\_load\_si512}\,($E_\theta[k, ..., k+15]$.node\_i); \\
      \hspace*{4em}$RJ:=$ \textbf{\_mm512\_load\_si512}\,($E_\theta[k, ..., k+15]$.node\_j); \\
      \hspace*{4em}// all following steps are performed SIMD-parallel\\
      \hspace*{4em}// on the 16 edges in $(RI, RJ)$ using AVX-512:\\
      \hspace*{4em}$RD:=$ Euclidean distances between nodes in $RI$ and $RJ$;\\
      \hspace*{6em}// determined by $2\cdot d$ operations ``\textbf{\_mm512\_i32gather\_ps}''\\
      \hspace*{4em}determine 16 parabolas $(d_{i,j}^+,w_{i,j}^+)$ acc. to Eq.~(\ref{eq:wplus}) and (\ref{eq:dplus});\\
      \hspace*{4em}perform a step of weighted majorization for the 16 edges;\\
      \hspace*{4em}$RG :=$ vectorizedSampleNonEdges\,$(RI)$; // cf. Fig~\ref{fig:nonedges}\\
      \hspace*{4em}determine $(d_{i,g}^-,w_{i,g}^-)$ and perform weighted majorization;\\
      \hspace*{4em}$RH :=$ vectorizedSampleNonEdges\,$(RJ)$; // cf. Fig~\ref{fig:nonedges}\\
      \hspace*{4em}determine $(d_{j,h}^-,w_{j,h}^-)$ and perform weighted majorization;\\
      \hspace*{4em}update $\mathbf y_{i}, \mathbf y_{j}, \mathbf y_{g}, \mathbf y_{h}, z_{i}, ...$ using gather and scatter;\\
      \hspace*{3em}explicitVectorizedReduction\,($\mathbf y_i, z_i$); // cf. Section~\ref{sec:reduction}\\
      \hspace*{2em}$\forall i\in\{0,...,n-1\}$: $\mathbf x_i := \frac{1}{z_i}\cdot \mathbf y_i;$\\
      \hspace*{2em}sample histograms and determine $\sigma$ according to Section~\ref{sec:slope};\vspace{0mm}\\
      \hspace*{1em}\textbf{return} $(\mathbf x_0, ..., \mathbf x_{n-1});$
    \end{minipage}
  }
  \caption{The Algorithm MulticoreGEMPE.}
  \label{fig:algo}
\end{figure}

\noindent Our overall algorithm is presented in Figure~\ref{fig:algo}. We apply MIMD-parallelism using OpenMP in the loop ``\textbf{foreach thread} $\theta$''. Each thread $\theta\in\{0,...,t-1\}$ obtains a contiguously stored subset of the edges of approximately the same size $m_\theta \approx m/t$, however taking into account to have a number of edges which is divisible by 16, and remaining chunks of 16 edges are distributed evenly over the threads.

Inside the loop ``\textbf{for} $k$'', we apply SIMD-parallelism using AVX-512: at each time of the algorithm, 16 edges are considered simultaneously. We fetch the indices of start- and end-vertices of the edges by \textbf{\_mm512\_load\_si512}-operations, since we assure that the end-vertices are stored in a separate array from the start-verties. Both arrays are aligned with cache lines for maximum performance. We fetch the coordinates $\mathbf x_i, \mathbf x_j$ of the previous round using gather-operations, and then compute distances, the parabolas, and the weighted majorization steps for these 16 edges, as well as 16 sampled non-edges with $RI$ and 16 sampled non-edges from $RJ$, all in parallel using 16-fold SIMD parallelism. Numerators and denominators of the considered vertices are finally updated using gather and scatter operations.

\section{Experimental Evaluation}\label{sec:exp}
\noindent \textbf{Comparison Methods and Parametrization.} We provide an implementation of MulticoreGEMPE at\\ \url{https://github.com/plantc/GEMPE}. We compare to the graph drawing methods MulMent \cite{DBLP:journals/tvcg/MeyerhenkeN018} for which we have obtained the C++ code from the authors. For visual comparison with classical graph drawing methods, we used the JUNG graph drawing API \url{http://jung.sourceforge.net/} which is written in Java. In effectiveness and runtime experiments, we compare to various graph representation learning methods for which optimized C++ code is available. As MulticoreGEMPE, these implementations make use of multicore MIMD parallelism as well as vectorized SIMD operations. We compare to node2vec \cite{DBLP:conf/kdd/GroverL16} \url{https://github.com/xgfs/node2vec-c}, DeepWalk \cite{DBLP:conf/kdd/PerozziAS14} \url{https://github.com/xgfs/deepwalk-c}, VERSE \cite{Tsitsulin:2018:VVG:3178876.3186120} \url{https://github.com/xgfs/verse} and LINE \cite{DBLP:conf/www/TangQWZYM15} \url{https://github.com/tangjianpku/LINE}. All comparison methods have been parameterized as recommended in the corresponding publications. MulticoreGEMPE does not require any input parameters. Only the number of threads needs to be specified as for all other methods supporting multicore parallelism. All experimental data sets are publicly available. All experiments have been performed on Intel Xeon Phi 7210 Knights Landing (KNL) with 1.3 GHz with 96 GB main memory and CentOS 7.4.1708 operating system. Whenever not otherwise specified, we used 8 cores.
\begin{figure*}[t]
\centering
\includegraphics[width=1.0\textwidth]{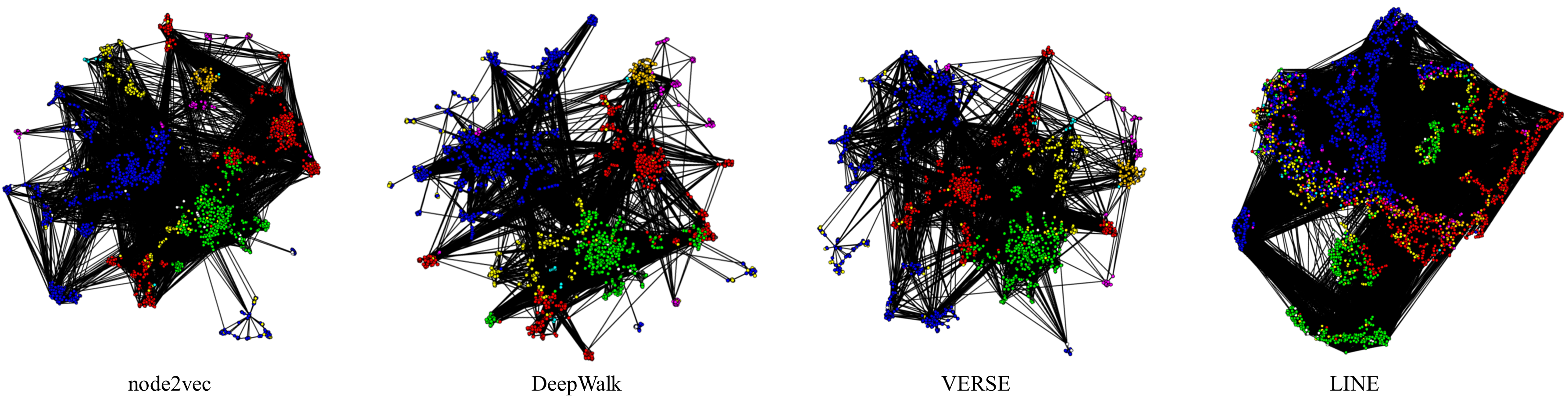}
\vspace{-1mm}
\caption{World Flight Network, comparison to graph representation learning methods. See also Figure \ref{fig:goodBad}.}
\label{fig:worldflights128}
\vspace{-2mm}
\end{figure*}

\subsection{Effectiveness}
\noindent \textbf{Graph Drawing.} The world flight network is available at \url{https://openflights.org/data.html}. The largest connected component consists of 3,304 nodes representing airports and 19,054 edges representing flights. For each airport, the data provides geo-coordinates and the assignment to large geospatial regions, e.g. the Americas, Europe and Asia, see Figure \ref{fig:goodBad}. Note that the world flight network is far from being a planar graph. There are many hubs connecting different continents by direct flights. Therefore, it is very difficult to obtain a 2D representation which matches the geo-coordinates just from the edges of the network. See Figure \ref{fig:goodBad} (top) for a drawing of the airports according to their geo-coordinates. There are a lot of edges especially connecting Europe and the Americas. Nevertheless, MulticoreGEMPE successfully captures the natural relationship of the continents just from the network information. For example, Africa in yellow color is located south of Europe (green), west of Asia (red) and east of America (blue). Australia (orange) is placed south of Asia.

Figure \ref{fig:worldflights128} shows drawings obtained from state-of-the-art graph representation learning methods. MulticoreGEMPE and these methods have been parameterized to generate 128D coordinates. As common practice in the representation learning literature, we used t-SNE \cite{ictdbid:2777} to map the resulting coordinates to 2D space. MulticoreGEMPE is the only method that preserves the natural relationship of the continents in 2D space, cf. Figure~\ref{fig:goodBad}, bottom. In the drawings of the other methods, these relationships are not visible. Node2vec and DeepWalk e.g. place America between Africa and Europe; In the drawing of VERSE, Australia is closer to most places in Africa than to most of Asia. The drawing of LINE does not show any natural relationship.

MulticoreGEMPE even preserves the shape of certain regions, especially North America and partly also Europe, Africa and Asia. In the drawing of node2vec, the shape of South America is visible. 
The coordinates of MulticoreGEMPE best fit the geo-coordinates with a normalized sum of squared errors (SSQ) of 0.63. Node2vec shows the second best with a SSQ of 0.71 followed by DeepWalk with 0.78, VERSE (0.81) and finally LINE with 0.82.

\begin{figure}[b]
\centering
\vspace{-3mm}
\includegraphics[width=1.0\columnwidth]{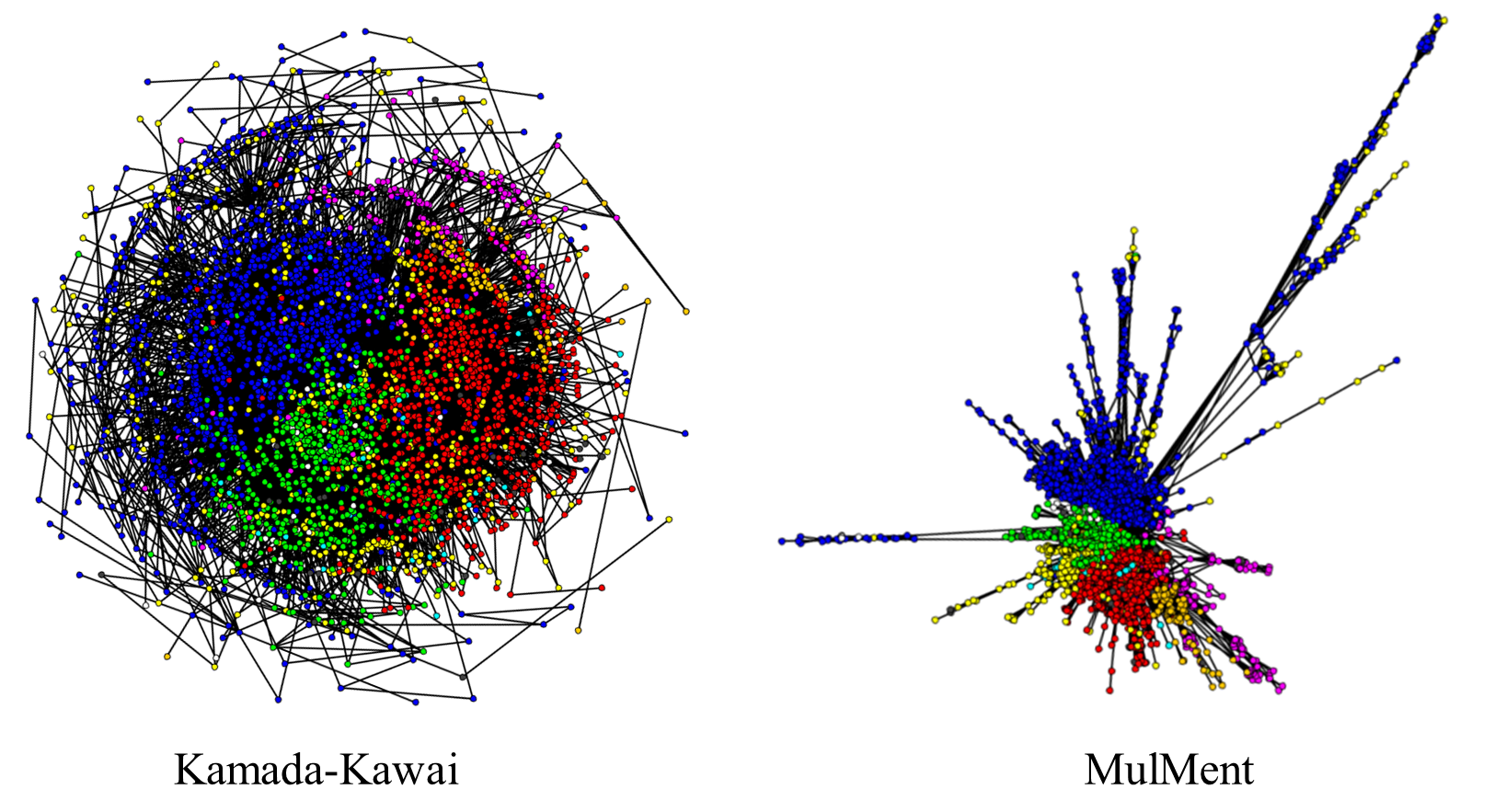}
\vspace{-2mm}
\caption{World Flight Network, comparison to graph drawing methods.}
\label{fig:worldflights2D}
\end{figure}

Figure \ref{fig:worldflights2D} shows the results of graph drawing methods which focus on extracting 2D coordinates. The classical Kamada-Kawai algorithm well separates the large continents but fails to preserve their relationships which yields an SSQ of 0.74. The Mulment algorithm also achieves 0.74. This drawing even more clearly separates the continents than Kamada-Kawai but does not make good usage of the drawing space as few nodes are spiked out. The self-organizing maps algorithm does not perform well with an SSQ of 0.94 (for space limitations not shown). In summary, we get best fit to the geo-coordinates when we first learn a high-dimensional representation with MulticoreGEMPE and then map it to 2D space. The geographical ground truth information is probably only weakly represented in the flight network. Similar to many social or biological networks, the world flight network is a small-world graph \cite{RePEc:eee:jaitra:v:14:y:2008:i:3:p:123-129}. Between any pair of cities, you need to take three or fewer flights. This small-world property dominates the geospatial relationships. The geographical information is included to some extent by the fact that small regional airports are connected to geographically close hubs. We need to extract higher dimensional coordinates to also preserve the geographical information.
\begin{figure*}[t]
\centering
\includegraphics[width=1.0\textwidth]{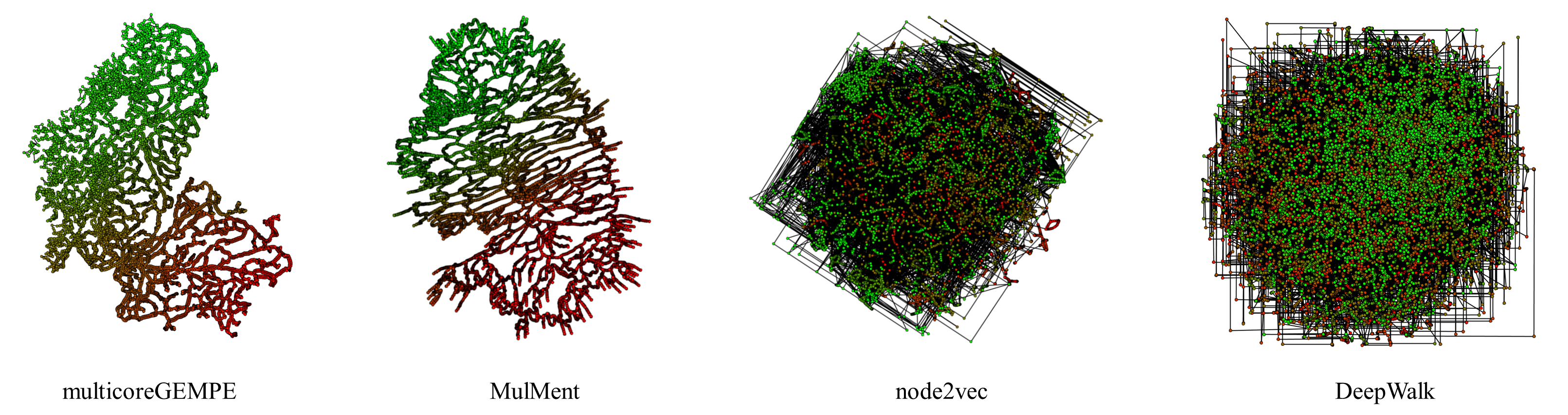}
\vspace{-3mm}
\caption{California road network. Nodes are colored from green (north) to red (south).}
\label{fig:california}
\vspace{-2mm}
\end{figure*}

Road networks can be much easier represented in 2D space. The node represent junctions and the edges represent road segments. There are no edges that directly connect remote places. Figure \ref{fig:california} compares the drawings of the California road network \cite{DBLP:conf/ssd/LiCHKT05} with 21,048 nodes and 21,693 edges. MulticoreGEMPE and the graph drawing technique MulMent successfully layouts the network with few edge crossings. The nodes are well aligned from north to south as visible by the transition from green to red color. For this network, we parameterized MulticoreGEMPE to generate 2D coordinates. As the other graph drawing methods, the implementation of MulMent only supports the generation of 2D coordinates. We cannot show the results of the Kamada-Kawai algorithm and self organizing maps as these methods do not finish within one day. The graph representation learning techniques are much more efficient. However, the results are not satisfactory. The shown results are obtained by generating 128D coordinates and mapping them down to 2D space with t-SNE as in the previous experiment. The result of VERSE is similar to that of DeepWalk and therefore omitted. LINE gives the worst result (not shown). We also tried to generate 2D coordinates directly but the results are even worse (not shown).

\noindent \textbf{Node Classification.} The major goal of graph representation learning is to extract a vector space representation from a graph in order to apply machine learning and data mining techniques. In order to compare the quality of the learned representations, we perform node classification experiments on graphs with node labels. The BlogCatalog \cite{Agarwal2020} data (10,312, nodes 333,983 edges) presents a very difficult classification task as nodes representing bloggers are labeled to 39 classes representing their interests. Table \ref{tab:blogcatalog} compares micro- and macro-F1-measure varying the amount of labeled nodes from 10\% to 90\%. Classification has been performed with a random forest classifier implemented in the Python \textit{scikit-learn} package. All experiments have been performed for 100 random splits into training and test data and we report the average result. MulticoreGEMPE clearly outperforms the comparison methods by a large margin. The best runner up is node2vec. The performance of MulticoreGEMPE on 10\% labeled nodes is better than the performance of node2vec on 90\% labeled nodes. The Cora and the PubMed data sets are citation networks available at \url{https://linqs.soe.ucsc.edu/data}. The Cora data (2,709 nodes, 5,429 edges) consists of publications labeled to 7 classes. On this network, all methods achieve much higher classification accuracy than on the BlogCatalog data with 39 classes. MulticoreGEMPE performs best for various amounts of labeled nodes. The best runner up is VERSE followed by DeepWalk, cf. Table \ref{tab:cora}. Also on the PubMed data (19,717 nodes 44,338 edges) with 3 classes, MulticoreGEMPE outperforms the comparison methods. Especially for few labeled nodes, the performance gains are substantial. The best runner ups are VERSE and node2vec, cf. Table \ref{tab:pubmed}. However, the comparison methods often require about twice the amount of training data to obtain the same performance. For instance, MulticoreGEMPE achieves a macro-F1-measure of 0.765 on 8\% of labeled nodes, while node2vec needs 16\% of labeled nodes to obtain 0.759 in macro-F1-measure.

%

\definecolor{lred}{rgb}{1.0, 0.5, 0.5}
\definecolor{dred}{rgb}{0.5, 0,0}
\begin{table}[b]
\begin{small}
\vspace*{-2mm}
\setlength{\tabcolsep}{1.5mm}
    \begin{tabular}{|l|r|r|r|r|r|r|}
        \hline& & & & & &\vspace{-3mm}\\
        \textbf{Macro-F1}& 10\% & 20\% & 30\% & 50\% & 70\% & 90\%\\\hline& & & & & &\vspace{-3mm}\\
        MulticoreGEMPE   & \color{dred}0.065 & \color{dred}0.078&\color{dred}0.084&\color{dred}0.095&\color{dred}0.097&\color{dred}0.107\\
        VERSE            & 0.004&0.007&0.008&0.010&0.013&0.013\\
        DeepWalk         & 0.010&0.014&0.016&0.020&0.021&0.022\\
        node2vec         & \color{lred}0.026&\color{lred}0.035&\color{lred}0.038&\color{lred}0.042&\color{lred}0.045&\color{lred}0.048\\
        LINE             & 0.001&0.000&0.000&0.000&0.000&0.000\\ \hline
    \end{tabular}\vspace{1mm}
    \begin{tabular}{|l|r|r|r|r|r|r|}
        \hline& & & & & &\vspace{-3mm}\\
        \textbf{Micro-F1}& 10\% & 20\% & 30\% & 50\% & 70\% & 90\%\\\hline& & & & & &\vspace{-3mm}\\
        MulticoreGEMPE   & \color{dred}0.159 & \color{dred}0.187 & \color{dred}0.196 & \color{dred}0.215 & \color{dred}0.221 & \color{dred}0.234\\
        VERSE            & 0.008 & 0.015 & 0.018 & 0.023 & 0.029 & 0.031\\
        DeepWalk         & 0.022 & 0.032 & 0.037 & 0.047 & 0.051 & 0.054\\
        node2vec         & \color{lred}0.065 & \color{lred}0.086 & \color{lred}0.092 & \color{lred}0.104 & \color{lred}0.112 & \color{lred}0.120\\
        LINE             & 0.002 & 0.001 & 0.001 & 0.001 & 0.001 & 0.001\\ \hline
    \end{tabular}
    \end{small}
    \vspace{1.5mm}
    \caption{Node Classification on BlogCatalog (\color{dred}Winner\color{black}/\color{lred}Runner-up\color{black}).}
    \label{tab:blogcatalog}
    \vspace{-2mm}
\end{table}

\begin{table}[b]
\begin{small}
\vspace*{-2mm}
\setlength{\tabcolsep}{1.5mm}
    \begin{tabular}{|l|r|r|r|r|r|r|}
        \hline& & & & & &\vspace{-3mm}\\
        \textbf{Macro-F1}& 0.5\% & 1\% & 2\% & 4\% & 8\% & 16\%\\\hline& & & & & &\vspace{-3mm}\\
        MulticoreGEMPE   & \color{dred}0.739 & \color{dred}0.758 &\color{dred}0.773 &\color{dred}0.783 &\color{dred}0.794&\color{dred}0.803\\
        VERSE            & \color{lred}0.633& \color{lred}0.684& \color{lred}0.721& \color{lred}0.744& \color{lred}0.760& 0.773\\
        DeepWalk         & 0.573& 0.636& 0.693& 0.731& 0.758& \color{lred}0.782\\
        node2vec         & 0.533& 0.615& 0.680& 0.723& 0.755& 0.780\\
        LINE             & 0.289& 0.289& 0.292& 0.294& 0.297& 0.298\\ \hline
    \end{tabular}\vspace{1mm}
    \begin{tabular}{|l|r|r|r|r|r|r|}
        \hline& & & & & &\vspace{-3mm}\\
        \textbf{Micro-F1}& 0.5\% & 1\% & 2\% & 4\% & 8\% & 16\%\\\hline& & & & & &\vspace{-3mm}\\
        MulticoreGEMPE   & \color{dred}0.760 & \color{dred}0.775 & \color{dred}0.788 & \color{dred}0.797 & \color{dred}0.807 & \color{dred}0.815\\
        VERSE            & \color{lred}0.697& \color{lred}0.730& \color{lred}0.754& \color{lred}0.770& \color{lred}0.783& 0.793\\
        DeepWalk         & 0.647& 0.692& 0.730& 0.759& 0.781& \color{lred}0.799\\
        node2vec         & 0.605& 0.669& 0.720& 0.753& 0.778& 0.798\\
        LINE             & 0.393& 0.395& 0.395& 0.396& 0.398& 0.399\\ \hline
    \end{tabular}
    \end{small}
    \vspace{1.5mm}
    \caption{Node Classification on Cora (\color{dred}Winner\color{black}/\color{lred}Runner-up\color{black}).}
    \label{tab:cora}
    \vspace*{-2mm}
\end{table}

\begin{table}[t]
\begin{small}
\vspace*{2mm}
\setlength{\tabcolsep}{1.5mm}
    \begin{tabular}{|l|r|r|r|r|r|r|}
        \hline& & & & & &\vspace{-3mm}\\
        \textbf{Macro-F1}& 0.5\% & 1\% & 2\% & 4\% & 8\% & 16\%\\\hline& & & & & &\vspace{-3mm}\\
        MulticoreGEMPE   & \color{dred}0.395 & \color{dred}0.538 &\color{dred}0.653 &\color{dred}0.724 &\color{dred}0.765&\color{dred}0.795\\
        VERSE            & 0.212 & \color{lred}0.350 & 0.464 & 0.573 & 0.682 & 0.741\\
        DeepWalk         & 0.184 & 0.288 & 0.396 & 0.531 & 0.658 & 0.738\\
        node2vec         & 0.212 & 0.313 &\color{lred} 0.467 & \color{lred}0.592 & \color{lred}0.689 & \color{lred}0.759\\
        LINE             & \color{lred}0.251 & 0.336 & 0.413 & 0.472 & 0.532 & 0.584\\ \hline
    \end{tabular}\vspace{1mm}
    \begin{tabular}{|l|r|r|r|r|r|r|}
        \hline& & & & & &\vspace{-3mm}\\
        \textbf{Micro-F1}& 0.5\% & 1\% & 2\% & 4\% & 8\% & 16\%\\\hline& & & & & &\vspace{-3mm}\\
        MulticoreGEMPE   & \color{dred}0.498 & \color{dred}0.617 & \color{dred}0.694 & \color{dred}0.745 & \color{dred}0.778 & \color{dred}0.806\\
        VERSE            & \color{lred}0.353 & \color{lred}0.474 & \color{lred}0.568 & 0.644 & 0.719 & 0.762\\
        DeepWalk         & 0.321 & 0.407 & 0.500 & 0.597 & 0.691 & 0.757\\
        node2vec         & 0.337 & 0.432 & 0.554 & \color{lred}0.648 & \color{lred}0.721 & \color{lred}0.777\\
        LINE             & 0.345 & 0.452 & 0.484 & 0.531 & 0.582 & 0.622\\ \hline
    \end{tabular}
    \end{small}
    \vspace{1.5mm}
    \caption{Node Classification on PubMed (\color{dred}Winner\color{black}/\color{lred}Runner-up\color{black}).}
    \label{tab:pubmed}
    \vspace{-3mm}
\end{table}

\subsection{Efficiency}
\begin{figure}[b]
\centering
\vspace*{-2mm}
\includegraphics[width=1.0\columnwidth]{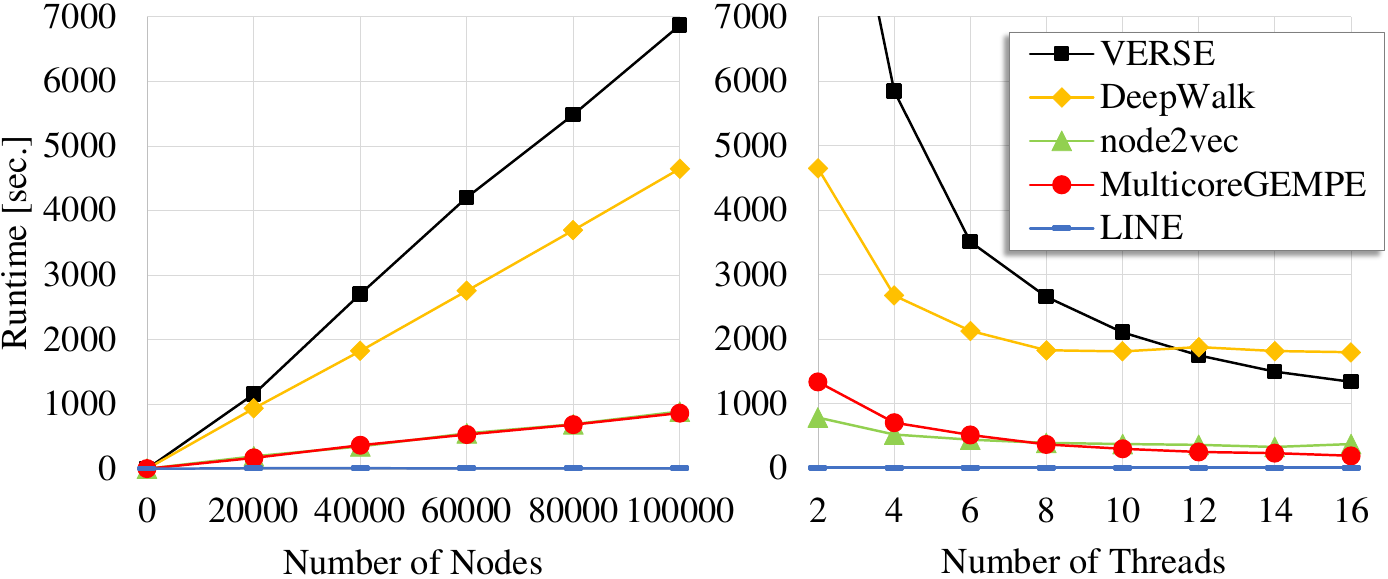}
\vspace{-2mm}
\caption{Runtime.}
\label{fig:runtime}
\end{figure}

\noindent For runtime and speedup experiments we compare MulticoreGEMPE to methods that are also capable to generate a representation of 128D and that are implemented in vectorized C++ code. Figure \ref{fig:runtime}(\emph{l.}) compares the runtime on synthetic graphs of various sizes which we have generated with the \textit{EppsteinPowerLawGenerator} implemented in the JUNG API. The number of edges increases linearly with the number of nodes. All methods have been run with 8 threads. All methods scale linearly with the number of edges. LINE is the fastest method, however also the worst in the effectiveness experiments. Node2vec and MulticoreGEMPE need approximately the same runtime. VERSE and DeepWalk need much more runtime than our method. Figure \ref{fig:runtime}(\emph{r.}) shows the speedup varying the number of threads on the data with 40,000 nodes. MulticoreGEMPE and (at an overall worse level) VERSE yield an almost linear speedup whereas node2vec and DeepWalk do not really profit much from increasing threads.

\section{Related Work and Discussion}
\label{sec:rw}
\subsection{High-performance Mining of Big Data}
\noindent We focus on scalable and effective graph representation learning on a commodity machine which is equipped with a multi-core microprocessor and a multi-level cache hierarchy. The GraphChi system \cite{DBLP:conf/osdi/KyrolaBG12} also focuses on a single machine setting but studies different graph mining problems, most importantly clustering of time evolving graphs. Related to our approach are recent techniques focussing on exploiting the interplay of multi-core MIMD parallelism with efficient SIMD instruction sets in current microprocessors for the efficient implementation of other algorithms e.g., k-means clustering \cite{DBLP:conf/sdm/BohmPP17}, similarity join in databases \cite{DBLP:conf/sigmod/PerdacherPB19}, Smith-Waterman \cite{DBLP:journals/ijpp/RucciSJGNP19} or the simulation of biomedical processes, e.g., \cite{DBLP:conf/iccS/JarvisLLNC19}. These approaches demonstrate that exploiting both levels of parallelism and thus making best usage of todays commodity machines requires completely re-engineering algorithms.

Other recent works focus on the usage of GPUs in order to speed up machine learning and data mining on graphs, see, e.g. \cite{DBLP:journals/pvldb/ChenDGP20, DBLP:conf/socinfo/ZhongH11, DBLP:conf/ipps/TangSPB20}. Many deep learning algorithms require a long training time for convincing performance. In this setting, a GPU or even a GPU cluster is anyhow mandatory. On a commodity machine, usually no powerful GPU is available and free for general purpose computations.

Many graph mining systems have been proposed for distributed environments, e.g., the PegasusN System \cite{DBLP:conf/aaai/ParkPK18}. It supports pattern mining and anomaly detection from very large graphs. Running under Spark, PegasusN can support larger graphs than the Pegasus System which is based on Map Reduce \cite{DBLP:books/crc/sakr14/Tsourakakis14}. These systems include some plotting functionality, e.g., spy plots highlighting specific patterns in the adjacency matrix, and distribution plots to spot outliers \cite{DBLP:conf/pakdd/KangLKF14}. However, these systems do not support graph representation learning, i.e. generating low-dimensional coordinates for the nodes of a graph that are suitable for data mining and visualization.

\subsection{Graph Drawing and Graph Representation Learning}
\noindent The task of finding a low-dimensional vector space representation for a graph has a long history and has been studied by different research communities. We categorize the variety of existing methods according to their different primary focus and the typical dimensionality of the extracted feature space. Methods from \emph{information visualization} aim at drawing graphs, for a survey see \cite{DBLP:journals/ivs/GibsonFV13}. Most related to MulticoreGEMPE from the family of info viz techniques are distance scaling methods which aim at mapping the shortest path distances to 2D space, e.g., \cite{DBLP:conf/gd/GansnerKN04,Freeman04graphictechniques,DBLP:journals/tvcg/MeyerhenkeN018}. Most of these techniques are based on the statistical technique of Stress Majorization which supports weights for individual node pairs. Most commonly, a weight of $w_{i,j} = d_{i,j}^{-2}$ is applied, where $d_{i,j}$ corresponds to the shortest-path distance between node $i$ and node $j$. Interestingly, this weighting is equivalent to the objective function of the classical Kamada Kawai algorithm. GEMPE exploits Majorization with a very different objective. Existing approaches consider distances and weights as input parameters. GEMPE iteratively learns target distances and weights optimizing the Predictive Entropy. We compare to MulMent \cite{DBLP:journals/tvcg/MeyerhenkeN018} which is a recent multi-level stress-based method specially designed for drawing large graphs. In contrast to the previously mentioned methods, MulMent supports graphs with several thousands of nodes like the California road map. We also compare to self organizing maps \cite{DBLP:conf/gd/Meyer98} as implemented in the ISOM layout of the JUNG API. As MulticoreGEMPE, this method does not require any input parameters, however it is outperformed by MulMent in our experiments. MulMent also outperformed the representation learning methods in graph drawing tasks, especially on inherently 2D networks, such as the California road network.\vspace{1mm}

Recent \emph{graph representation learning techniques} (for a survey see \cite{DBLP:journals/debu/HamiltonYL17}) focus on feature extraction from graphs. The methods DeepWalk  \cite{DBLP:conf/kdd/PerozziAS14} and node2vec \cite{DBLP:conf/kdd/GroverL16} learn the embedding of a node based on random walks. These methods optimize a cross-entropy loss between the probability of a node pair to co-occur in a random walk and the Euclidean distance in vector space.  In our experiments, DeepWalk and node2vec perform well in node classification. The technique LINE \cite{DBLP:conf/www/TangQWZYM15} considers preserving first-order and second-order node proximity, i.e. embeds nodes close to each other which are connected by an edge or have similar neighbors.  In most experiments LINE has been outperformed by other methods, most importantly VERSE and AROPE. VERSE \cite{Tsitsulin:2018:VVG:3178876.3186120} is based on the idea that an expressive graph embedding should capture some similarity measure among nodes. As a default, the Personalized Pagerank similarity is preserved. VERSE produced very good coordinates for node classification. A lot of further techniques with more specialized goals have been proposed. Struc2vec \cite{DBLP:conf/kdd/RibeiroSF17} and role2vec \cite{DBLP:journals/corr/abs-1802-02896}, e.g., focus on preserving the structural roles of the nodes in the embedding.
Some recent methods use autoencoders to represent the local neighborhood of nodes, such as SDNE \cite{DBLP:conf/kdd/WangC016}. AROPE \cite{DBLP:conf/kdd/ZhangCWPY018} exploits an eigen-decompositon reweighting theorem to reveal node relationships across several orders of proximity. GEMPE preserves arbitrary order proximities by its self-adapting distance scaling strategy guided by the Predictive Entropy. We introduced related information-theoretic objective functions for other graph mining tasks, such clustering \cite{DBLP:conf/icdm/GoeblTBP16, DBLP:conf/sdm/MuellerHSPB11}, summarization \cite{DBLP:conf/kdd/FengHKBP12}, link prediction \cite{DBLP:conf/icdm/FengHHBP13} and for dimensionality reduction of general metric data \cite{DBLP:conf/icdm/Plant14}.

\section{Conclusion}
\label{sec:conclusion} \noindent We have introduced MulticoreGEMPE, an efficient algorithm for graph drawing and graph representation learning on a commodity machine equipped with a multi-core CPU. Our algorithm is inspired by the idea to construct a vector space representation of a graph that compresses the link information in the adjacency matrix. MulticoreGEMPE is designed to exploit the MIMD paralellism and the SIMD parallelism of current microarchitectures. Our experiments demonstrate that MulticoreGEMPE is highly competitive with the state-of-the-art in graph drawing and graph representation learning. Unlike most existing methods, MulitcoreGEMPE produces very good results in graph drawing and representation learning on various types of networks ranging from inherently 2D road networks to inherently very high-dimensional social networks. 
\section*{Acknowledgment}
\noindent \small{This work has been funded by the German Federal Ministry of Education
and Research (BMBF) under Grant No. 01IS18036A. The authors of this work take full responsibility for its content.}
\nocite{mci/Boehm2009,DBLP:conf/sigmod/PerdacherPB19}\bibliographystyle{plain}
\bibliography{bibliographyNew}
\end{document}